# Real-Time Vehicle Detection And Urban Traffic Behavior Analysis Based On Uav Traffic Videos On Mobile Devices

Yuan Zhu [a] Yanqiang Wang [a] Yadong An [a] Hong Yang [b] Yiming Pan [a]

[a] *Inner Mongolia Center for Transportation Research, Inner Mongolia University, Hohhot, China*
[b] *Department of Computational Modeling and Simulation Engineering, Old Dominion University, Norfolk, America*

**A B S T R A C T**

This paper focuses on a real-time vehicle detection and urban traffic behavior analysis system based on Unmanned Aerial Vehicle (UAV) traffic video. By using UAV to collect traffic data and combining the YOLOv8 model and SORT tracking algorithm, the object detection and tracking functions are implemented on the iOS mobile platform. For the problem of traffic data acquisition and analysis, the dynamic computing method is used to process the performance in real time and calculate the micro and macro traffic parameters of the vehicles, and real-time traffic behavior analysis is conducted and visualized. The experiment results reveals that the vehicle object detection can reach 98.27% precision rate and 87.93% recall rate, and the real-time processing capacity is stable at 30 frames per seconds. This work integrates drone technology, iOS development, and deep learning techniques to integrate traffic video acquisition, object detection, object tracking, and traffic behavior analysis functions on mobile devices. It provides new possibilities for lightweight traffic information collection and data analysis, and offers innovative solutions to improve the efficiency of analyzing road traffic conditions and addressing transportation issues for transportation authorities.

## 1. Introduction

In recent years, with the increasing demand for traffic safety and congestion mitigation tasks, real-time traffic data collection has gained significant attention from researchers. However, traditional traffic data collection methods, including manual collection and equipment collection such as photodetectors, ultrasonic detectors, require substantial labor and material resources, which cannot meet the current demand for high accuracy, timeliness and predictability of traffic data analysis and prediction. Modern methods such as traffic electronic monitoring system collection, GPS, mobile phone, GIS and other information data collection methods have shortcomings such as fixed angle position and poor timeliness. Therefore, the use of UAVs for traffic video collection is more convenient than traditional methods, which is an effective means to solve the problem of traffic information collection and can ensure the real-time nature of traffic data.

Mobile devices are portable and flexible, make it capable to collect traffic data anytime and anywhere. Although the computing power of mobile devices is far inferior to that of server-side, the performance of Apple's mobile silicon and the machine learning frameworks provide hardware and software support for object recognition and trajectory tracking on the mobile platform.

At present, most of the research on real-time detection and tracking through UAV video is based on theories by improving the algorithms and assessing the performance on datasets, which lacks practical applications. In addition, there are several factors to be considered for different UAVs and mobile devices.

---

Based on the existing theoretical research, this paper presents practical experiments that integrate UAVs, machine learning, and mobile development technology. It proposes a system based on the iOS platform, which controls UAVs to capture real-time video streams, perform object detection and tracking of vehicles at intersections. The system analyzes and visualizes the detection results to obtain micro and macro traffic parameters. The paper describes the program's design process and conducts experiments to validate its feasibility. Through this research, this paper expects to improve the real-time and accuracy of traffic data collection in practical applications and provide support for solving urban traffic problems.

## 2. Literature review

The flexibility, ease of operation and real-time of aerial video make it a useful tool in the data acquisition of urban traffic. Zhang et al. *(1)* utilized fixed UAV angle to capture traffic videos and employed the Mask-RCNN model for vehicle detection. The pixel distance between the center point of the vehicle's Bounding Box in fixed interval time frames was calculated and converted to actual distance. By analyzing the vehicle speeds, the model was able to determine whether traffic congestion occurred and identify the causes of congestion. Byun et al. *(2)* proposed a method that utilizes EfficientDet for vehicle detection and SORT for tracking to analyze the movement of vehicles. By segmenting the road area in the drone image and calculating the ratio of lane length to pixels, the actual speed of the vehicle can be determined. Chalmers et al. *(3)* employed frame sampling technology in conjunction with DJI Mavic Pro 2 to achieve real-time object detection of animals, maximizing both accuracy and throughput.

Recent advancements in traffic trajectory recognition using drones and computer vision have shown significant progress on desktop-side devices. Gu et al. *(4)* proposed a framework for analyzing collision risk in highway interchanges by scrutinizing drone-captured video data based on vehicle micro-behavior. Chen et al. *(5)* utilized drone data to observe mixed traffic flow involving motorized vehicles, non-motorized vehicles, and pedestrians within safety space boundaries. Wu et al. *(6)* developed an automatic road conflict identification system (ARCIS) using the masked region convolutional neural network (R-CNN) technique to process traffic videos collected by UAVs. Ma et al. *(7)* focused on using UAV vehicle trajectory data to establish a traffic conflict prediction model for highway traffic in diversion areas. Chen et al. *(8)* achieved high-precision trajectory extraction and fast, accurate vehicle tracking by UAVs using kernel technology and coordinate change, combined with wavelet transform. The Intelligent and Safe Transportation Laboratory (UCF SST) at the University of Central Florida employed deep learning techniques to extract parameters from UAV videos, facilitating road safety diagnosis *(9)*. Additionally, they created the CitySim dataset *(10)* to support traffic safety analysis and digital twin scene modeling.

While most studies have conducted performance evaluations on datasets after improving the models, theoretically meeting the requirements for real-time detection performance, practical applications of these methods have not been fully tested yet *(11-15)*. From the standpoint of the model deployment platform, Hua et al. *(16)* introduced a lightweight UAV real-time object tracking algorithm based on policy gradient and attention mechanism, and successfully deployed the model on NVIDIA Jetson AGX Xavier devices. In related research, several studies have utilized the Nvidia Jetson TX2 on-board computer for object detection and tracking computations *(17, 18)*. Regarding the utilization of mobile devices, Martinez-Alpiste et al. *(19)* conducted a series of tests using OpenCV in conjunction with the YOLOv3 model deployed on an Android phone. They evaluated the feasibility of achieving machine learning-based object detection on mobile platforms, employing frames per second, accuracy, battery consumption, temperature, RAM usage, model size, FLOPS, and model load time as evaluation indicators. In the context of iOS-based systems, Zhou et al. *(20)* executed the object tracker on the iPad Air2, while Li et al. *(21)* deployed a YOLO model trained on the COCO dataset on the iPhone to achieve object detection functionality. However, some challenges were encountered, such as bounding box shifts.

The problem in existing methods is the image-based detection or tracking methods typically require huge computational power and complicated algorithms which make it hard to implement. In this paper, we first propose a novel method that can effectively tune the thresholds in detection algorithms. This paper takes advantage of the flexibility of the modern mobile devices and deep learning based real-time object detection method to realize a mobile system which can automatically detect certain objects from the video stream of UAV.

## 3. Mobile-based object detection and tracking system

The system's core design is centered on the iOS framework, aiming to seamlessly integrate drone technology. Customized drone functionalities are developed to facilitate real-time control of drone flight and video frame data acquisition. This integration enables simultaneous target detection during video frame acquisition, resulting in real-time visualization of detected vehicle targets. The system's ultimate goal is to harmoniously merge UAV flight control, traffic video acquisition, target detection, trajectory tracking, and data analysis on a mobile terminal."

*3.1 Overall structural design of the system*

The system is predicated on the iOS Software Development Kit (SDK), which serves as the foundational framework for formulating the core system functionalities and interface design. Building upon this foundation, the DJI SDK is leveraged to actualize custom drone control functionalities, enabling the real-time acquisition of the drone's video stream. In tandem, machine learning frameworks including Core ML and Vision are deployed on the iOS platform to facilitate real-time processing of the video stream. This processing is chiefly focused on the detection and tracking of vehicle targets. For an in-depth visualization of the system's structure and components, please refer to Fig. 1.

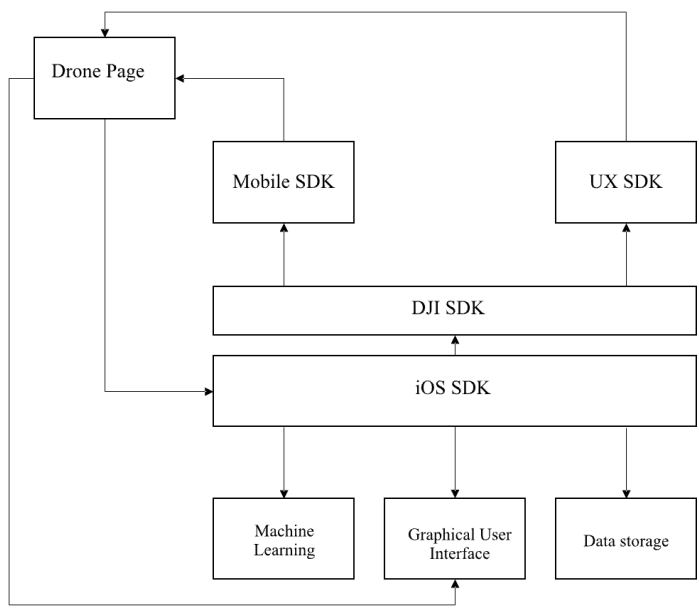

**Fig. 1.** Overall structure of the system

The system comprises four distinct sub-functions, each with its dedicated interface. These sub-functional modules encompass the following capabilities: (1) real-time vehicle detection and analysis, drawing upon UAV traffic video; (2) vehicle detection and analysis using locally stored UAV traffic video; (3) retrospective analysis based on detection data; and (4) retrieval of flight logs and data export.

As shown in Fig. 2, the real-time vehicle detection and analysis function, relying on UAV traffic video, integrates video frame acquisition, vehicle target detection and tracking, traffic behavior analysis, and data visualization. The encapsulation of each function within independent modules serves to enhance system scalability and maintainability, facilitating the decoupling of functionalities. This design strategy ensures heightened adaptability and efficiency in responding to evolving requirements during future development and maintenance.

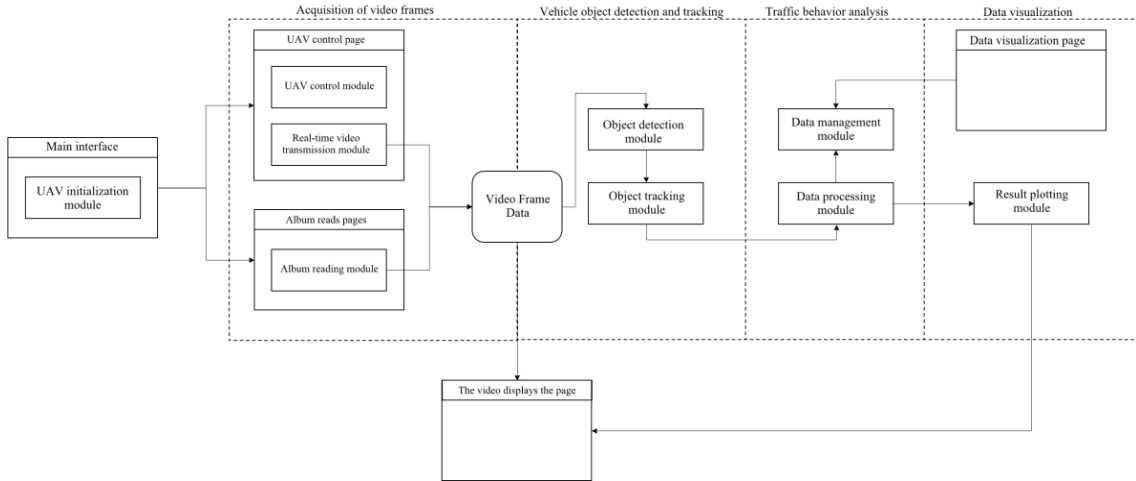

**Fig. 2.** Implementation of vehicle detection and analysis

*3.2 Design and functional implementation of the main interface*

For enhanced management of both the primary function interface and sub-function interfaces, this system employs the Navigation Controller as its foundational controller. Upon system initialization, the user is directed to the main interface. The main interface is structured in accordance with the Model-View-Controller (MVC) design pattern, leveraging the UIKit framework for user interface (UI) design. Its primary responsibilities encompass system initialization and drone setup.

(1) Interface design
The interface is partitioned into left and right sections. The left area serves as the functional zone, housing 6 UIButtons to facilitate functions such as data review, viewing mobile phone albums, accessing flight records, checking binding and activation statuses, and logging into DJI accounts. Additionally, 4 UILabels are employed to display binding status, activation status, connection mode, and the currently connected device. The right section is dedicated to establishing the connection with drones through a single UIButton. The interface layout, as illustrated in Fig. 3, reflects this division.

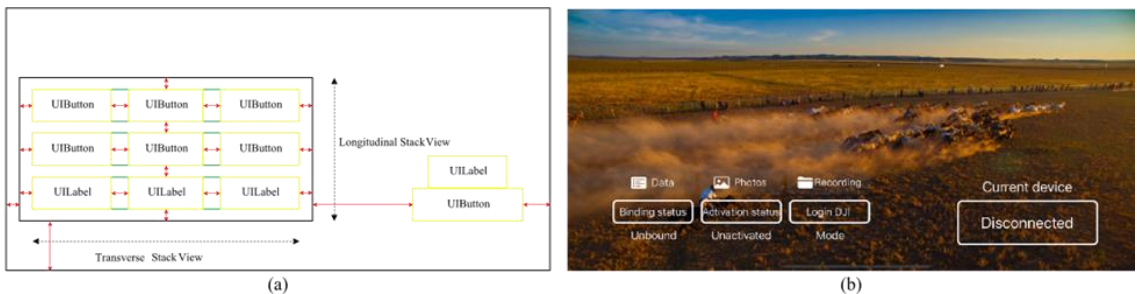

**Fig. 3.** Graphical User Interface layout design (iPhone device)

5To ensure consistent styling of UI controls with similar functionalities, Outlet Collections are employed for synchronous management. Given the presence of multiple UIButtons and UILabels within the functional area, and considering the system's extensibility, a Stack View is utilized to encapsulate and manage the design. The functional area is structured as a unit within the Stack View.

(2) Function implementation

In the realm of main interface business logic, the Model layer assumes responsibility for the data management module, while the Controller layer is overseen by DJIViewController. The View layer's controls are linked to DJIViewController through the outlet mode and managed via the attribute mode. A method connection is established with DJIViewController in response to control-triggered events, employing the action mode. The correspondence between controls and their functionalities is delineated in Table 1.

By clicking the relevant control, the segue scene transition method of the Show type is triggered. This action pushes the corresponding function's view controller into the stack of the current navigation view controller, effectuating the interface transition. It is imperative to note that the initiation of drone control necessitates validation logic from the drone initialization module before activation.

Table 1 DJIViewController control attribute table

| Control | Type | Property | Method | Function |
| --- | --- | --- | --- | --- |
| Recording | UIButton | djiFileButton | showDjiFlyFile | Get a flight log |
| Data | UIButton | reviewButton | reviewData | View historical probe data |
| Photos | UIButton | mobileAlbumButton | getMobileAssets | View the album |
| Binding state | UIButton | bindButton | bindingState | Query the binding status |
| Activation state | UIButton | activeButton | activateState | Query the activation status |
| Login Dji | UIButton | loginDJIButton | loginDJIAccount | Log in to your DJI account |
| Unbound | UILabel | aircraftBindingStateLabel | / | Displays binding status |
| Unactivated | UILabel | activateStateLabel | / | Displays activation status |
| Mode | UILabel | modeLabel | / | Displays connection mode |
| Disconnected | UIButton | connectDJIButton | connectDJI | Jump to the probe interface |

*3.3 UAV video frame acquisition scheme*

The real-time image transmission module assumes the critical role of receiving the live video stream generated by the drone. This stream is subsequently relayed to the detection and tracking module, which, in turn, transmits the video stream data to the vehicle. The real-time image transmission module serves as a pivotal intermediary, directly bridging the UAV's real-time video stream to the target detection module. In doing so, it provides a vital interface for seamless interaction with the target detection module.

(1) Interface design

The design layout of the real-time video transmission interface is depicted in Fig.4 , delineating two distinct areas within a two-dimensional plane perspective: the upper and lower segments.

The upper area serves as the status information section, designated for monitoring the drone's status information. This upper area is further subdivided into three distinct sections: the UAV information area, PTZ (Pan-Tilt-Zoom) information area, and other information area. These sub-areas are visually represented through the utilization of UILabels and are structured using a horizontal Stack View. The UAV information area displays real-time processing performance metrics, including the UAV's pitch, roll, and yaw angles. The PTZ information area provides details on the gimbal's pitch, roll, and yaw angles. Additionally, the other information area presents essential data such as the current time, drone boot time, and the number of satellites.

The lower area is designated as the image transmission display zone, denoted as FPV View. This area is constructed using the root view UIView of the FPVViewController. From a three-dimensional perspective, a custom UIView layer is superimposed onto the underlying view. This additional layer is employed to create an area dedicated to displaying detection results, aptly named Detect View. The purpose of this Detect View is to visually represent the outcomes of target detection.The implementation specifics of the drone interface are graphically illustrated in Fig. 5.

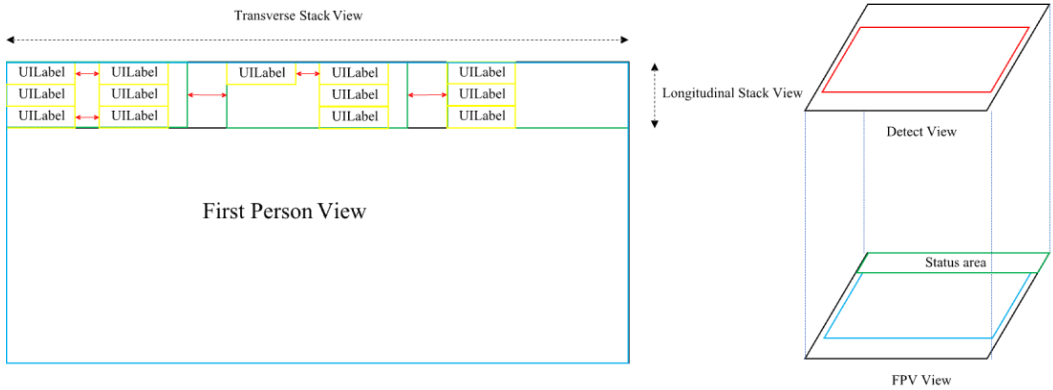

**Fig. 4.** Live video module horizontal and vertical design diagrams

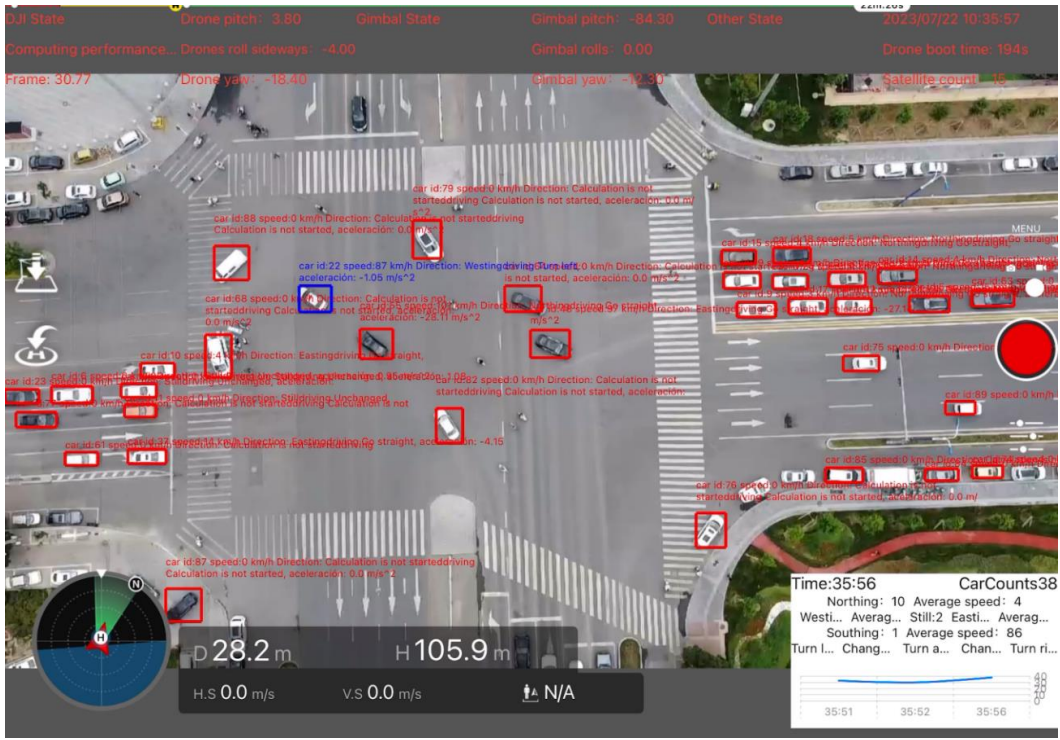

**Fig. 5.** UAV control interface implementation effect (iPad device)

(2) The overall implementation of the real-time image transmission module

In the context of the real-time image transmission module, the implementation is characterized by the distribution of responsibilities across the Model, Controller, and View layers.

The Model layer takes charge of the data management module, handling the processing and management of data. The Controller layer is overseen by the FPVViewController class, which serves as the central controller for the module. Meanwhile, the View layer establishes a property connection with the FPVViewController, employing the outlet mode to manage controls as properties. To facilitate seamless interaction and event-driven responses, a method connection is established with the FPVViewController through the action mode, triggered by specific control events. The relationship between controls and their corresponding functionalities is detailed in Table 2.

Table 2 FPVViewController control attribute list

| Control | Type | Property | Function |
| --- | --- | --- | --- |
| FPS | UILabel | fpsLabel | Displays the current video frame |
| UAV Pitch | UILabel | djiPitchLabel | Displays the pitch angle of the drone |
| UAV Roll | UILabel | djiRollLabel | Displays the roll angle of the drone |
| UAV Yaw | UILabel | djiYawLabel | Displays the yaw angle of the drone |
| Gimbal Pitch | UILabel | gimbalPitchLabel | Displays the gimbal pitch angle |
| Gimbal Roll | UILabel | gimbalRollLabel | Displays the gimbal roll angle |
| Gimbal Yaw | UILabel | gimbalYawLabel | Displays the yaw angle of the gimbal |
| Current Time | UILabel | currentTimeLabel | Displays the current time |
| Starting Duration | UILabel | djiStartTimeLabel | Displays the drone start-up time |
| Satellite Quantity | UILabel | satelliteCountLabel | Displays the number of satellites connected |
| Start | UIButton | startDetectButton | Start identifying and tracking |
| Detect View | DetectView | detectView | Plot probe results in real time |
| FPV View | UIView | fpvView | Real-time display of the video transmission interface |

As illustrated in Fig. 6, the drone's video stream pertains to the transport stream of video data, which is conveyed to the mobile device employing the drone's camera as the transmission medium. Typically, due to the constraints imposed by image transmission bandwidth limitations, the original video stream necessitates compression before it can be transmitted to the mobile device. Consequently, encoding on the drone's side and subsequent decoding on the mobile device's side become indispensable. In the case of DJI drones, the encoding technology is encapsulated within DJI's proprietary video transmission systems, such as OcuSync and Lightbridge.

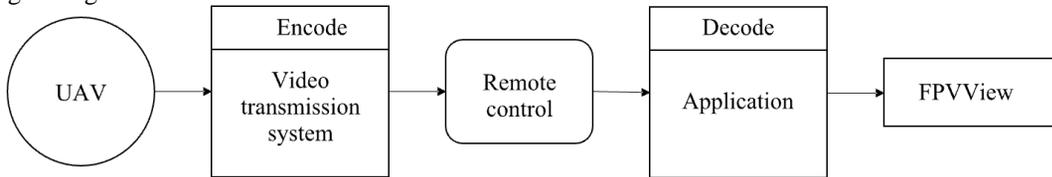

**Fig. 6.** Real-time live video data flow link

The core function of the real-time image transmission module revolves around the presentation of the drone's live video feed. Concurrently, it plays a pivotal role in forwarding the video frame data to both the target detection and target tracking modules. Subsequently, the analysis results, furnished by the data processing module, are instantaneously rendered on the screen through the result drawing module. The communication structure that orchestrates this information flow is meticulously depicted in Fig. 7.

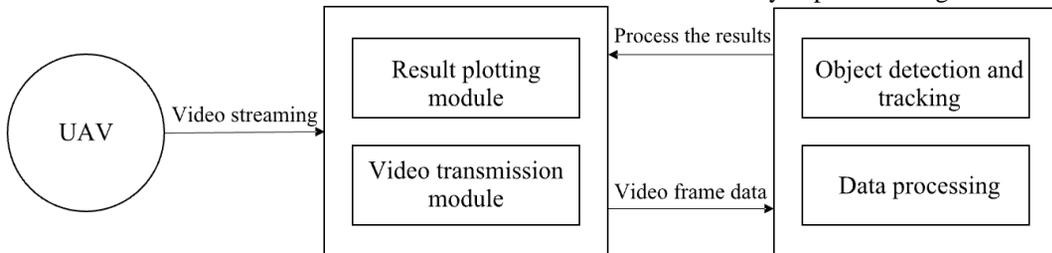

**Fig. 7.** Connection diagram of real-time detection function submodules

Addressing the real-time challenge of target detection necessitates careful consideration of the mobile terminal's computational capabilities for video frame processing and the acquisition of detection results. Given that the video stream is acquired in real-time from an aerial drone camera, the UAV video frame transmission rate to the system remains consistently at 30 frames per second (FPS). When the tasks of video retrieval, recognition, and result rendering are executed sequentially, inadequacies in processing capacity can lead to a backlog of unprocessed pre-sequence frames, while post-sequence frames continue to queue for processing. This, in turn, results in task blockages, causing a failure to process and display video frames promptly. Such inefficiencies may further amplify memory usage, increase CPU load, and eventually lead to system crashes.

To address this challenge, the real-time video transmission module employs two intermediary methods to process the video stream. It delineates the separation of real-time video stream display and result processing from the drawing process. Furthermore, the module incorporates a dual-layered view structure along the Z-axis direction, enhancing its capacity to manage real-time image transmission efficiently.

The foundational view, FPV View, is entrusted with the responsibility of rendering the real-time video stream. Simultaneously, the top-tier view, Detect View, undertakes the task of visualizing the detection results. These two layers of views are meticulously superimposed to ensure that the system's video frame processing rate remains unaffected by the image transmission rate. When the rate of video frame processing surpasses the video transmission rate, this harmonious integration facilitates real-time detection. In situations where the video frame processing rate is equal to or slower than the video transmission rate, a frame loss policy may be employed to align the processing rate with the transmission rate, ensuring the consistency of data processing. The processing and execution of the video stream are detailed in Fig. 8

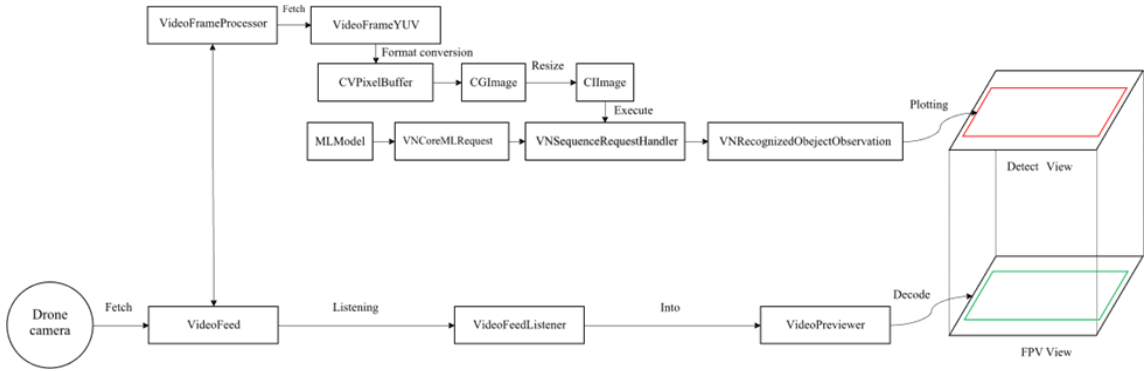

**Fig. 8.** Schematic diagram of implementing real-time detection dual proxy method

The realization of real-time image transmission is a structured process. To begin, we override the lifecycle method "viewDidLoad()" within the FPVViewController view controller. In this method, the singleton object of the decoder, DJIVideoPreviewer, is acquired. Subsequently, the root view of FPVViewController is designated as the target object for the decoder via the `setView()` method. Furthermore, the properties "enableHardwareDecode" and "enableFastUpload" of the decoder are configured to enable hardware-based decoding and rapid GPU-based uploading. In the next phase, the "viewWillAppear" lifecycle method is redefined. This method is executed just before the view becomes visible. Here, we utilize the 'videoFeeder()' method from DJISDKManager to obtain the video provider object, DJIVideoFeeder. Notably, some DJI drones are equipped with multiple cameras, allowing for the transmission of up to two camera data streams simultaneously. However, the drones employed in this study are single-camera drones. Therefore, the primary video source is obtained through DJIVideoFeeder's "primaryVideoFeed". We establish a listener method to be implemented by the current controller and initiate the decoding process by invoking the "start()"method.

Finally, within the callback method "videoFeed:didUpdateVideoData()" of the listener agent DJIVideoFeedListener, the video stream data, encapsulated in the "videoData"object, is retrieved and subsequently relayed to the decoder through the "push()" method.

The live video streaming framework in this documentation encompasses key elements: the video provider object, VideoFeed, the video stream listener agent, DJIVideoFeedListener, the decoder object, DJIVideoPreview, and the display interface, FPV View. This interconnected framework is visually represented in the data flow diagram, as depicted in Fig. 9.

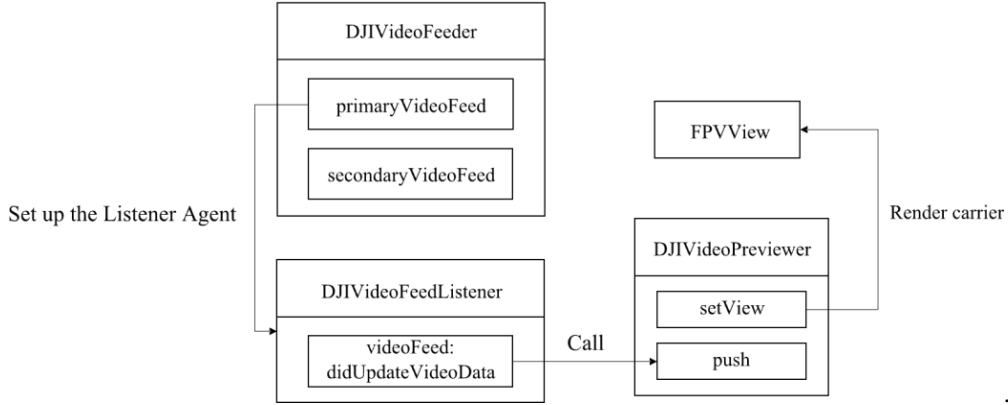

**Fig. 9.** Video data flow direction

In light of the inability to directly transmit video stream data obtained through the DJIVideoFeedListener listening agent to the object detection module, the real-time video transmission module employs the VideoFrameProcessor proxy method to acquire video frame data. The VideoFrameProcessor's proxy method is invoked with each retrieval of a video frame. In the callback method "videoProcessFrame()", the VideoFrameProcessor retrieves the video frame data object, denoted as "frame". This object, of type VideoFrameYUV, is subject to a condition check to determine the initiation of the probe, contingent upon the evaluation of whether "isStartDetect" is true. Subsequently, the VideoFrameYUV data frame undergoes a transformation to a CVPixelBuffer type through the "createPixelBuffer()" method. This CVPixelBuffer is then relayed to the target detection module and the target tracking module for data frame processing. The target tracking module returns an array of detection result rectangles, denoted as "rects". This information is employed to calculate micro-level parameters and macro-level traffic parameters of the vehicle within the data processing module.

To visualize these detection results, the "showInDetectView()" method is utilized. It passes the processing result rectangles to the "polyRects" property of the custom view, "detectView". Subsequently, the "draw()" method of the "detectView" is triggered, leading to the interface's recalibration on the main queue, accomplished via the "setNeedsLayout()" method of the "detectView".

Furthermore, the "viewWillDisappear" lifecycle method is overridden to facilitate memory management. When the back button is clicked, returning to the main interface, several critical actions are undertaken. These include the removal of the target view of the DJIVideoPreviewer decoder through the "unSetView()" method, the elimination of the DJIVideoFeedListener through the "remove()" method, thereby halting the reception of new video frame data. Finally, the "close()" method is invoked to shut down the decoder, releasing system memory resources.

*3.4 Vehicle target detection and tracking implementation Tracking module based on SORT algorithm*

The current implementation of the Vision tracker exhibits certain performance bottlenecks in object tracking. To address this, we propose an enhancement by replacing the Vision framework tracker with a method based on the SORT algorithm. The improved tracking module encompasses the following seven key steps.

(1) The object detection module transmits the detection results, which are then received and processed for each object's bounding box properties. To ensure compatibility with the SORT tracker, the y-axis of the bounding box is flipped, and the coordinates are adjusted from the Vision framework to the UIKit framework. Normalized coordinates are mapped to real image coordinates, and the format is converted from [x, y, w, h] to [x1, y1, x2, y2].

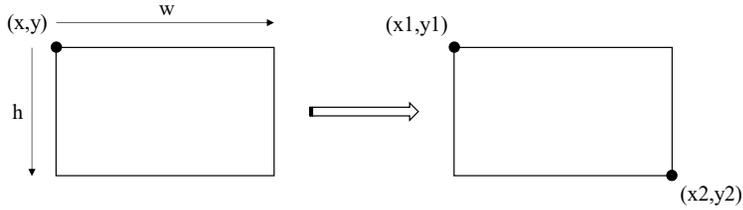

**Fig. 10.** Coordinate conversion

(2) Each detected object is assigned a unique ID, and a Kalman filter is set up to predict their state in the next frame, initializing the tracker.

(3) The new detections and existing trackers undergo data correlation using IoU matching. The Hungarian algorithm calculates the IoU distance and identifies the best matches, associating newly detected objects with the corresponding existing trackers.

(4) For objects that have successful matches, their Kalman filters are updated based on observation results, leading to improved tracking accuracy.

(5) Object tracking is achieved by updating their position information according to the predicted state of the Kalman filter. Temporary occluded objects have a specified life cycle, allowing them to remain "lost" for a certain duration. If no matching object is found within this period, the corresponding tracker is removed.

(6) New trackers are created for newly detected objects that do not match with existing ones, and their Kalman filters are initialized.

(7) The tracking results are generated by providing status information (e.g., ID, location). Iterating through the results returned by the tracker, the format is converted from [x1, y1, x2, y2] to [x, y, w, h]. Additionally, the detected object are updated with the relevant ID assigned by the tracker.

## 4 Analysis and visualization of traffic behavior based on detection data

The target tracking module delivers instantaneous positional data. Situated between the target tracking module and the result drawing module, the data processing module assumes the pivotal role of processing every frame of data forwarded by the target tracking module. Within this data processing paradigm, the module undertakes the computation of real-time micro-level parameters for individual vehicles while simultaneously deducing macro-level traffic parameters with respect to lanes or road segments. Subsequently, these computed parameters find their path to the result plotting module for visual representation. This architectural arrangement materializes an efficient amalgamation of data processing and analysis, resulting in a noteworthy enhancement of precision and practicality in the domain of traffic behavior analysis.

*4.1 Calculation of sampling intervals and scale bars*

In the context of real-time data processing, the employment of fixed-frame sampling introduces variability in the time intervals between samples. This variance adversely impacts the calculation of traffic micro parameters, leading to deviations and, consequently, imprecise speed calculations. Conversely, if a single frame is selected for sampling, the resulting interval proves excessively brief. Such brevity introduces inaccuracies stemming from the detection accuracy error of the target's position and errors in the time interval calculations. The outcome is pronounced fluctuations in speed calculation results.

To address these challenges, this study adopts a dynamic calculation approach. By computing the average FPS, this method utilizes the FPS value as the foundation for determining the sampling interval in the acquisition of traffic data. This strategy effectively mitigates the influence of errors on the results by adjusting and extending the sampling interval.

(1) Calculation of sampling intervals

To initiate the probing process, each activation of the "Start Probing" button triggers the creation of a temporary variable, "frames," designed to tally the cumulative number of frames per probe. This variable is initialized with a value of 0 and is iteratively incremented by 1 during the processing of each individual frame. This mechanism ensures an accurate count of the frames captured during each probing operation.

Subsequently, the processing duration of each frame is meticulously recorded by computing the timestamps both before and after the processing of a single frame. The collective processing time $timeAll$ is determinable using Eq. (1).

$$timeAll = \sum_{i=1}^{n} endTime(i) - startTime(i) \tag{1}$$

Where *startTime* and *endTime* are the timestamps before and after single frame processing.

Conclusively, the real-time processing frame rate $FPS$ is derived from the calculation presented in Eq. (2).

$$FPS = \frac{frames}{timeAll} \tag{2}$$

Where $frames$ is number of frames during each detection.

As illustrated in Fig. 11, the data of FPS video frames serves as the basis for defining the sampling period interval. By configuring the sampling interval coefficient, with the default value of 1 representing a one-second interval, the sampling interval size becomes dynamically adaptable. The actual calculated sampling interval $SFPS$ is ascertained using Eq. (3).

$$SFPS = FPS * timeRatio \tag{3}$$

Where *timeRatio* is the sampling interval coefficient.

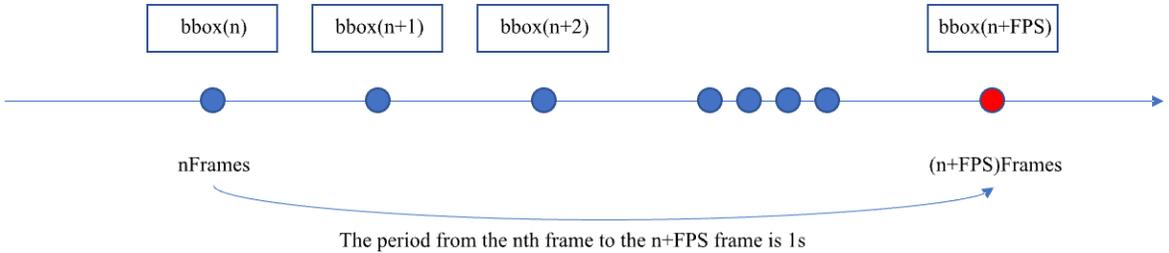

**Fig. 11.** Sampling interval schematic

(2) Calculation of sampling intervals

As shown in Fig. 12, the use of drones in various settings introduces variations in lens angles and target sizes, which are contingent upon drone models and different flight altitudes. In the context of real-time object detection and tracking, measurements of target size are conventionally provided in screen pixel coordinates. Therefore, in the process of calculating traffic-related data, it is imperative to establish a conversion factor relating pixel distances to actual distances.

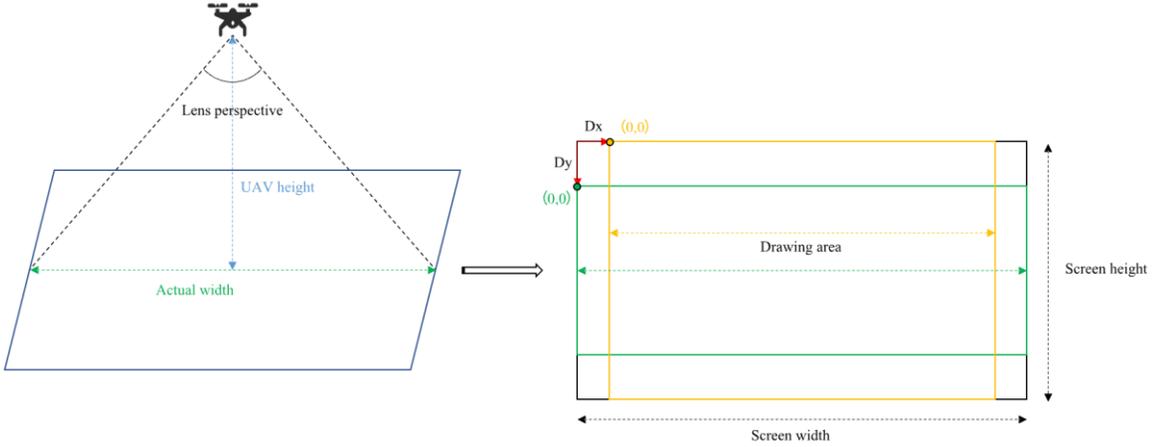

**Fig. 12.** Comparison of actual drone images and device images

First, the height height of the drone gets the altitude property of the DJIFlightControllerState class instance from the callback method by implementing the DJIFlightControllerDelegate proxy delegate in the drone module.

First, the altitude (referred to as "height") of the drone is determined by accessing the "altitude" property of the DJIFlightControllerState class instance. This information is obtained through the implementation of the DJIFlightControllerDelegate proxy delegate within the drone module.

Subsequently, the lens angleof the drone is assigned multiple values corresponding to various drone models via an enumeration structure. The drone module, in turn, ascertains the specific drone model by employing an agent and matching it with the associated lens angle. This approach enables the calculation of the actual width $realWeight$ of objects captured by the drone, as per Eq.(4).

$$realWeight = 2 * height * \tan\frac{angle}{2} \quad (4)$$

Where *angle* is the lens angle of the drone, *height* is the altitude of the drone.

In the final step, the dimensions of the plotting area (referred to as "imageAreaRect") are determined through the drawing module. The "imageAreaRect" encapsulates the pixel width and height of the display area, facilitating the calculation of the real-world distance represented by each pixel referred to as *pixelToRealDistance*, as outlined in Eq. (5).

$$pixelToRealDistance = \frac{realWeight}{imageAreaRect.weight} \quad (5)$$

Where $imageAreaRect.weight$ is the the plotting area, *weight* is the altitude of the drone.

In summary, after obtaining the distance to the pixel, it can be mapped to the real distance for traffic data calculation.

*4.2 Calculation of traffic parameters*

Micro parameters are calculated in frames after receiving the trace results for each frame, and the data is inserted into the database through the data management module.

(1) Calculation of speed
First, the vehicle target moves between two frames as shown in Fig. 13.

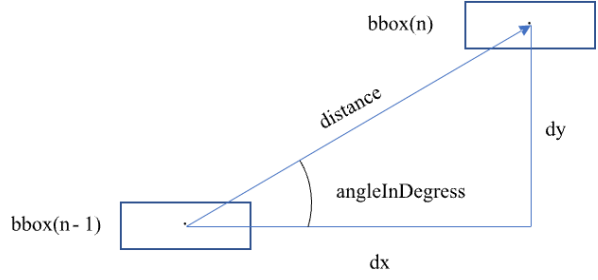

**Fig. 13.** Schematic of vehicle target's movement distance in one frame

The pixel distance moved by the vehicle target for one frame can be calculated by Eq. (6)-(8).

$$dx = bbox(n) \cdot midX - bbox(n-1) \cdot midX \qquad (6)$$

$$dy = bbox(n) \cdot midY - bbox(n-1) \cdot midY \qquad (7)$$

$$distancePixel(n) = \sqrt[2]{(dx)^2 + (dy)^2} \qquad (8)$$

where *bbox* is the target box of the vehicle, *midX* and *midY* are the horizontal and vertical coordinates of the center, *dx* and *dy* are the horizontal and vertical offsets of the center between each frame, and *distancePixel* is the distance in pixels that the vehicle moves in 1 frame.

Second, the distance traveled by the vehicle target in a calculation cycle is shown in Fig. 14.

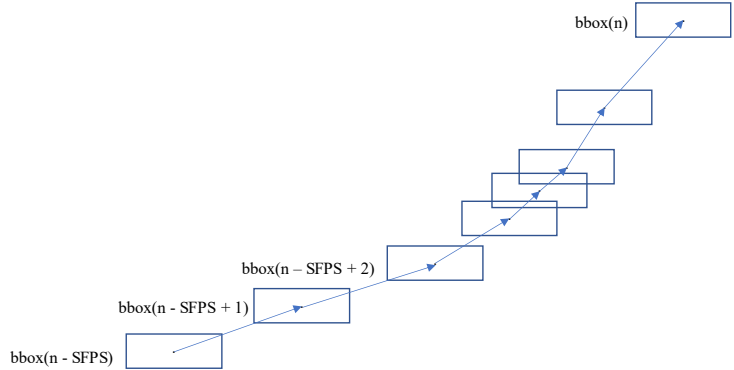

**Fig. 14.** Schematic of actual movement distance of vehicle target in the sampling interval

The route of the vehicle will change during the moving process. There will be an error, if the Euclidean distance between the two endpoints is directly calculated, so the exact moving distance is obtained by calculating the sampling interval through Eq. (9).

$$realDistance(n) = \sum_{i=n-SFPS}^{n} distancePixel * pixelToRealDistance \qquad (9)$$

where *realDistance* is the actual distance traveled by the vehicle, *SFPS* is the sampling interval, *pixelToRealDistance* is the pixel distance ratio.

Finally, the vehicle moving speed is calculated by Eq. (10).

$$speed(n) = \frac{realDistance(n)}{timeRatio} \qquad (10)$$

where *timeRatio* is the sampling interval coefficient, *speed* is the actual speed of the vehicle.

(2) Calculation of acceleration

Take the speed parameters of the two endpoints of the sampling interval and calculate the acceleration by Eq. (11).

$$acceleration(n) = \frac{speed(n) - speed(n - SFPS)}{timeRatio} \quad (11)$$

where *acceleration* is the acceleration of a vehicle.

(3) Judgment of the direction of movement

First, the current vehicle speed is judged, and if the vehicle speed is zero, the vehicle is stationary. Second, when the vehicle speed is not 0, calculate the offset of the vehicle target in the x-axis and y-axis directions in one sampling interval, and calculate the displacement angle by Eq. (12).

$$angleInDegress = \frac{atan2(dy, dx) * 180}{\pi} \quad (12)$$

where the *atan2* function represents the relative angle of the line between the start point and the end point and the positive x-axis, so that the *angleInDegress* represents the angle with the positive direction of the x-axis, and the interval is between (-180°, 180°).

Finally, the moving direction of the vehicle can be obtained by the interval where the displacement angle is located. As shown in Fig. 15, the lane change judgment can be made after obtaining the direction of movement.

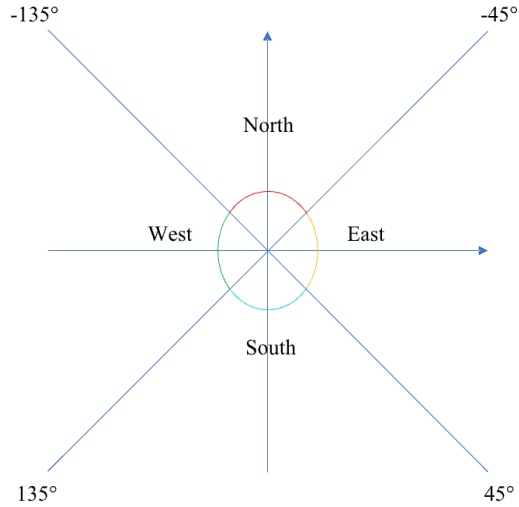

**Fig. 15.** Schematic of vehicle displacement angle and movement direction correspondence

(4) Lane change judgment

The Euclidean distance between sampling points is calculated by Eq. (13) and compared with the lane change thresholds.

$$isLaneChange = \begin{cases} true, distance > Threshold \\ false, distance \leq Threshold \end{cases} \quad (13)$$

wherer *isLaneChange* is the lane change identifier, *distance* is the Euclidean distance between the sampling points.

When the conditions for lane changing are met, the directions of the two end points of the sampling interval are obtained, the state before and after them is judged to calculate whether lane changing or steering occurs, and finally the lane changing behavior in each direction is judged by the positive and negative values of *dx* and *dy*. The calculation of driving north is shown in Fig. 16.

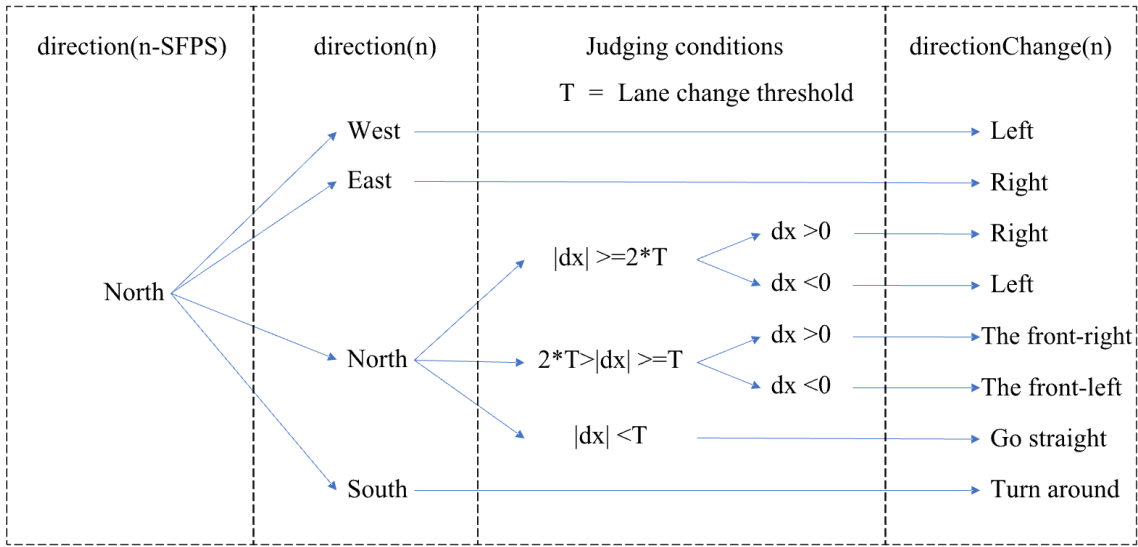

**Fig. 16.** Diagram of lane change for northbound traffi

The statistical analysis of macro-parameters is carried out following the calculation of micro-parameters, with the current detection results counted at regular intervals of every FPS frame. In this study, a rectangular box can be drawn at the center of the intersection by sliding on the screen, automatically generating a total of eight lane areas as illustrated in Fig. 17, representing east-west, south-north directions.

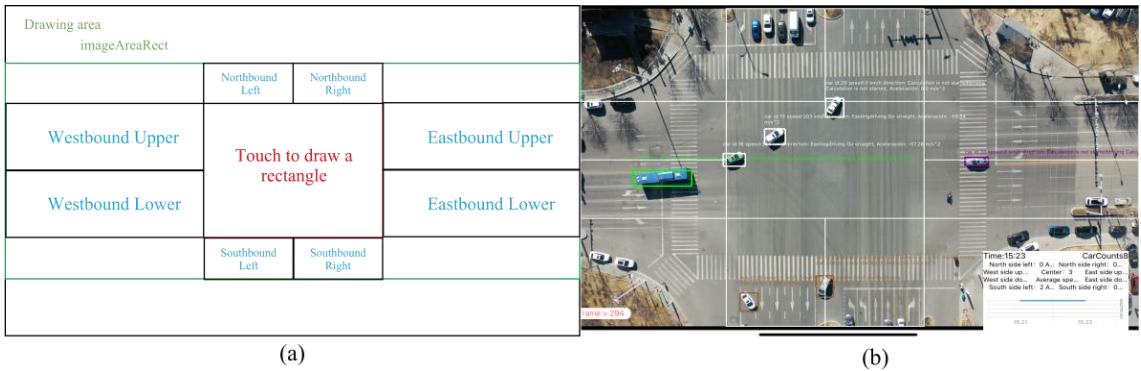

**Fig. 17.** Lane area schematic

Subsequently, the center point location of each vehicle is matched with the corresponding lane area in the detected vehicles. By doing so, the number of cars in each lane area can be accurately counted. This process enables us to obtain vital information concerning the distribution of vehicles in different lanes, thereby facilitating better traffic congestion monitoring, road traffic direction planning, and traffic diversion strategies.

*4.3 Data Visualization Module*

Real-time object detection and tracking are implemented through two distinct views, with the result plotting module responsible for rendering the vehicle target frames and micro traffic data. Additionally, the data visualization interface extracts macro traffic data from a database and offers visual representation. When processing results are conveyed to the result drawing module, it triggers the setNeedsLayout() method of DetectView, which, in turn, invokes the draw() method. This process leverages the Core Graphics framework to craft the target results. The drawing process is segmented into four fundamental stages: drawing context configuration, drawing area computation, target box rendering, and target information box rendering.

(1) Drawing Context Configuration: To configure the drawing context, the current graphics context "ctx" is acquired using the "UIGraphicsGetCurrentContext()" method. It is noteworthy that the coordinate system origin within the Core Graphics frame is situated in the lower left corner. Upon the return of the "draw()" method, the system automatically aligns the coordinate system with the UIKit coordinate system. The "saveGState()" method is employed to create a copy of the current graphics state, which is pushed to the top of the context graphics state stack. Subsequently, the context is configured to act as a transparent artboard, ensuring that the drawing layer does not obstruct the FPV View's video transmission screen.

(2) Computation of the Drawing Area: The video stream resolutions vary among different UAV models, while screen sizes differ across various iOS devices. Consequently, the scaling ratio of the video frame displayed on the screen exhibits variations, as exemplified in Fig.18. Importantly, the drone's video stream does not occupy the entire screen; hence, it becomes necessary to compute the dimensions of the drawing area to enable accurate rendering of the target boxes.

The aspect ratio of the video stream is determined using the pre-acquired property "captureDeviceResolution." Subsequently, the size of the graphic drawing area, relevant to the current device's screen size, is computed and represented as "imageAreaRect." It's crucial to note that the iOS system automatically scales the video frame to fit the screen. In this regard, two scenarios arise during video frame scaling: "case 1," involving alterations in height based on width, and "case 2," leading to changes in width based on height. The calculations are as follows:

The aspect ratio $AspectRatio$ of the video stream is calculated as per Eq. (14).

$$AspectRatio = \frac{width}{height} \tag{6}$$

Where *width* and *height* are the resolution of the video stream.

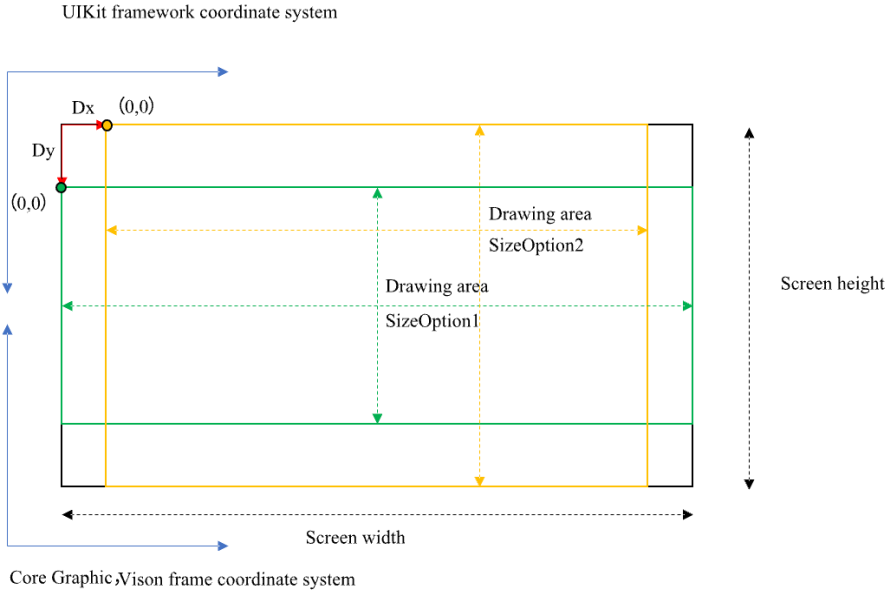

**Fig. 18.** Drawing area and screen area schematic

To commence, a determination is made to identify whether the scenario corresponds to case 1. The process entails the calculation of the scaled area's dimensions through Eq. (15), represented as *SizeOption1*. Subsequently, an evaluation ensues to ascertain whether the height of *SizeOption1* is less than or equal to the height of the "DetectView "view. The fulfillment of this condition confirms that it pertains to case 1.

$$SizeOption1 = (V.width, \left\lfloor \frac{V.width}{AspectRatio} \right\rfloor) \quad (7)$$

Where *V.width* and *V.height* are the size of the Detect View.

The computation of the *Dx* and *Dy* offsets for the drawing area is accomplished by applying Eq. (16) and Eq. (17), while the dimensions of *SizeOption1* are employed as the dimensions of the *imageAreaRect* within the drawing area.

$$Dx = \begin{cases} 0, & SizeOption1.height \leq V.height \\ \left\lfloor \frac{V.width - SizeOption2.width}{2} \right\rfloor, & SizeOption1.height > V.height \end{cases} \quad (8)$$

$$Dy = \begin{cases} \left\lfloor \frac{V.height - SizeOption1.height}{2} \right\rfloor, & SizeOption1.height \leq V.height \\ 0, & SizeOption1.height > V.height \end{cases} \quad (9)$$

Conversely, if the earlier condition is not satisfied, the situation is identified as case 2. In this context, the calculation of the scaled area's dimensions transpires through Eq. (18), denoted as *SizeOption 2*. Concurrently, the offsets *Dx* and *Dy* are determined through Eq. (16) and Eq. (17). The dimensions of *SizeOption2* are employed as the dimensions of the drawing area.

$$SizeOption2 = (\lfloor V.heigth * AspectRatio \rfloor, V.height) \quad (10)$$

(3) The process of target box rendering involves the traversal of the result array "polyRects" designated for rendering. For each target object "polyRect" within the array, distinct drawing colors are assigned based on the current state of the vehicle. The "cornerPoints" attribute of the polyRect object is obtained, providing the coordinates of inflection points. An adjustment to the coordinate system origin of these inflection point coordinates is executed, aligning it with the coordinate system origin of the imageAreaRect within the drawing area. Subsequently, the inflection points are traversed and connected through the utilization of the "addLine()" method, culminating in the closure of the target box.

(4) Proceeding to the drawing of the target information box, the size of the text box is derived from the "boundingBox" property of the polyRect object. An offset is applied to the "textRect" to align it with the imageAreaRect coordinate system through the use of the affine transformation "CGAffineTransform". The offset positions the text box 1 height unit above the rectangular box while retaining the same height adjustment. The width of the text box is established as 4 times the size of the rectangular box. The text information is formatted in rich text style, incorporating the micro parameters of the vehicle target. This information is drawn within the text box area.

Lastly, the graphical context artboard is preserved in its most recent state using the "restoreGState()" method. This concludes the draw method, culminating in the rendering of the detection results for the current frame and the refresh of the detectView interface.

## 5. Model Training and Experiments

This chapter elaborates on the training and deployment procedures of the object detection model, encompassing comparative experiments designed to scrutinize the varying performance and efficacy of each model. To assess the practicality of real-time target detection and analysis, a series of comparative experiments were undertaken to validate the system's operational effectiveness. These experiments encompass the training of distinct models, feasibility assessments, real-time target detection performance validation, real-time target tracking performance validation, erformance testing, and real-world flight experiments.

*5.1 Dataset selection and processing*

A total of three datasets were employed in the experimental phase, with the primary dataset originating from a multitude of intersection traffic videos captured by DJI drones within the urban confines of Hohhot, designated as the Hohhot dataset. To enhance the dataset's diversity, this study amalgamated the VisDrone dataset, the UAVDT dataset, and the Drone Vehicle dataset.

(1) Hohhot dataset: This dataset encompasses vehicle videos recorded at various prominent intersections within the urban region of Hohhot, as depicted in subfigure (a) of Fig. 19. The videos were acquired under varying altitudes, lighting conditions, weather circumstances, and across multiple temporal segments, including morning and evening rush hours. The drone captures footage from a 90° vertical downward perspective.

(2) VisDrone dataset: As illustrated in subfigure (b), this dataset comprises videos from 14 distinct cities, encompassing diverse vehicle densities, weather conditions, and environmental settings. The dataset encompasses ten target categories, ranging from diminutive targets such as pedestrians and bicycles. The VisDrone dataset comprises 400 video clips, comprising 265,228 frames and 10,209 static images, all captured via drones.

(3) UAVDT dataset: Displayed in subfigure (c), this dataset exhibits a resolution of 1080*540, recorded at 30FPS, encompassing video frame sequences under differing weather conditions, altitudes, angles, and levels of occlusion. The dataset revolves around three primary target categories: CAR, BUS, and TRUCK.

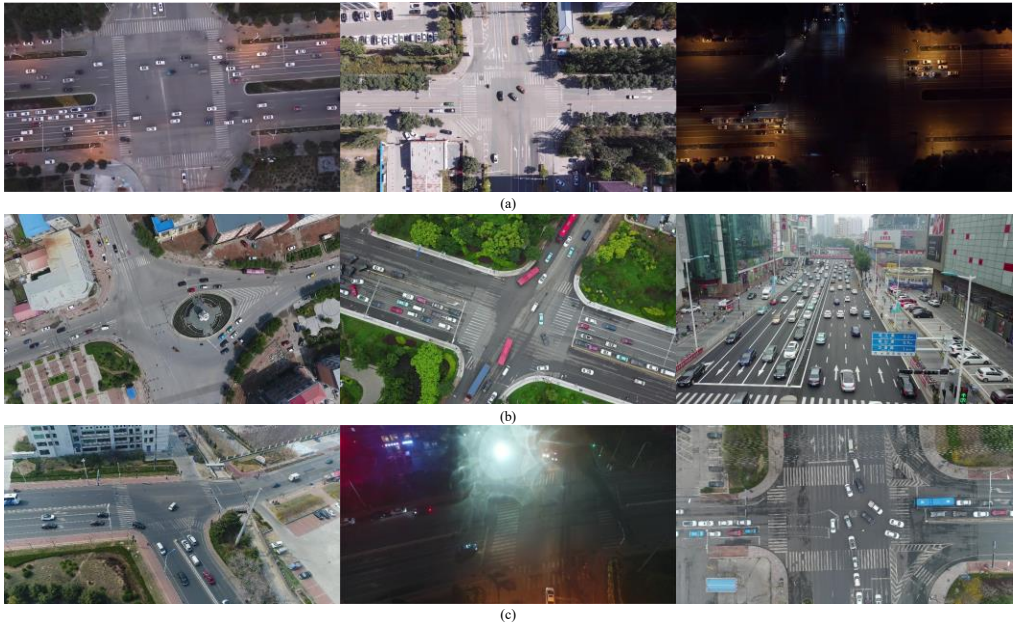

**Fig. 19.** Main dataset

Through the amalgamation of these three datasets, the primary objective of this study is to create a composite dataset characterized by diverse scenes, categories, and shooting conditions. This comprehensive dataset is intended to facilitate a more precise evaluation of the object detection and tracking algorithm's real-world performance. Concurrently, this amalgamation serves to enhance the model's capacity for generalization, enabling it to deliver robust detection results across various scenarios.

Preprocessing of the Hohhot dataset: The initial step involves extracting video frames from the Hohhot dataset, with each video frame having a resolution of 1920*1080 pixels. Subsequently, the ImageLabel tool is deployed to annotate the vehicles in these frames, with cars, buses, and trucks being uniformly labeled as "cars." Annotation tools are employed to rectify or exclude invalid data and to ensure a balanced distribution of labels across the dataset. Moreover, data augmentation techniques are applied to enhance the overall dataset diversity.

Preprocessing of third-party datasets: The original UAVDT, VisDrone, and Drone Vehicle datasets undergo preprocessing using Python tools. This process involves filtering label files to retain only the data captured from a vertical perspective. Subsequently, the data is categorized into three classes based on shooting height: "high," "medium," and "low." The data is further refined to encompass only three target categories: "car," "bus," and "truck," which are subsequently standardized as "car." Redundant information is pruned, leaving behind essential details such as the target category and bounding box coordinates, represented as label, x, y, w, and h.

Dataset fusion and partitioning: Following the consolidation of the three datasets, the resolution is standardized to 640 x 640 pixels, and the dataset is partitioned into multiple sub-datasets. Additionally, the label information is converted into TXT format, employing normalized coordinates, suitable for the YOLO algorithm, and JSON format, utilizing pixel coordinates, which is intended for use with Create ML. Through this comprehensive sequence of preprocessing steps, the dataset's quality and diversity are meticulously curated, thereby establishing a robust foundation for subsequent model training and performance assessment.

*5.2 Model Training Target tracking performance testing*

During the model training process, two distinct methods were employed: utilizing the Create ML tool for training the native model and employing Coremltools to convert the YOLO model based on the PyTorch framework.

(1) Training based on Create ML

In the Create ML training process, the verification set partitioning method was set to automatic (auto). Create ML automatically determined the proportion of the verification set based on the dataset size. Various batch-size settings were explored, including 16, 32, 64, 128, and auto, resulting in 56 training iterations. Through observation of the model's performance under different parameter configurations, adjustments were made to optimize the model. The best-performing cubic model was selected from these iterations, as presented in Table 3.

Table 3 Three model data tables based on Create ML training

| training number | Loss | Training set accuracy (%) | Validation set accuracy (%) | Number of iterations | Batch-Size |
|---|---|---|---|---|---|
| train_1 | 5.2 | 71% | 44% | 19000 | auto |
| train_8_2 | 2.377 | 93% | 92% | 10000 | 16 |
| train_182 | 0.971 | 95% | 96% | 10000 | 16 |

In all training experiments, the longest successful training duration was 28.5 hours, while failed training lasted up to 38 hours. The optimal training effect was achieved with a batch size of 16 and 10,000 iterations. Increasing the dataset size and iteration count elevated the failure rate. Larger batch sizes led to proportionate increases in training time. In unsuccessful training attempts, the optimal Loss value converged to 2.32.

(2) Training based on YOLO model

The dataset was divided into training, validation, and test sets in a 7:2:1 ratio. The training environment specifications are outlined in Table 5.9. Utilizing the PyTorch framework, three YOLO models, namely YOLOv5, YOLOv7, and YOLOv8, were trained. The models' characteristics facilitated their conversion into mobile models. The model conversion algorithm was modified, incorporating the NMS module function. This modification allowed for direct model visualization on macOS and enabled the filtering of target detection results through IOU and confidence level. Separating bounding box functionality from business logic enhanced system efficiency and facilitated model usage on iOS platforms.

Training for YOLOv5 involved utilizing the YOLOv5m initial weights, with an input size of 640*640 and a batch size of 4. Training proceeded through three distinct phases: 5, 50, and 500 epochs. Fig. 21 illustrates the results after 5 epochs of training.

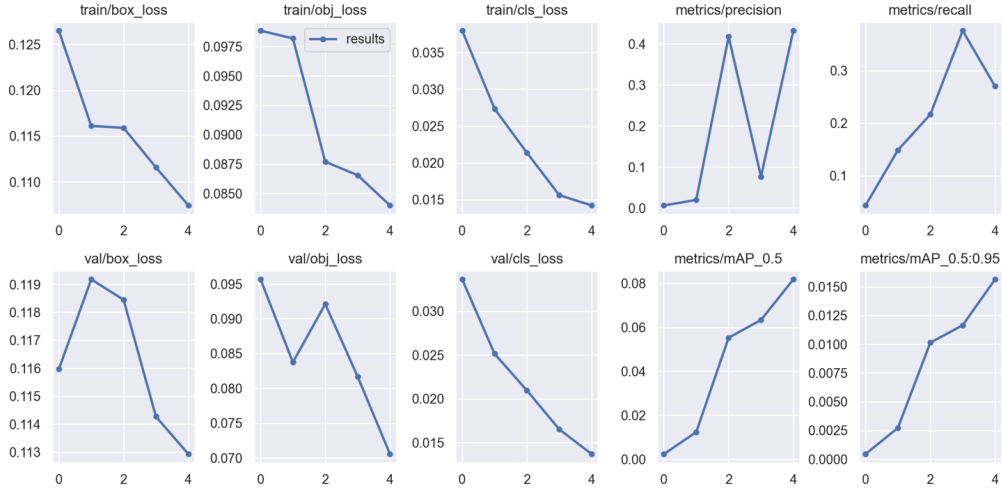

**Fig. 20.** YOLOv5 training results

For YOLOv7, the training employed the YOLOv7 initial weights, an input size of 640*640, and a batch size of 4. The training regimen included three phases: 5, 50, and 500 epochs. The outcomes after 500 epochs are depicted in Fig. 22.

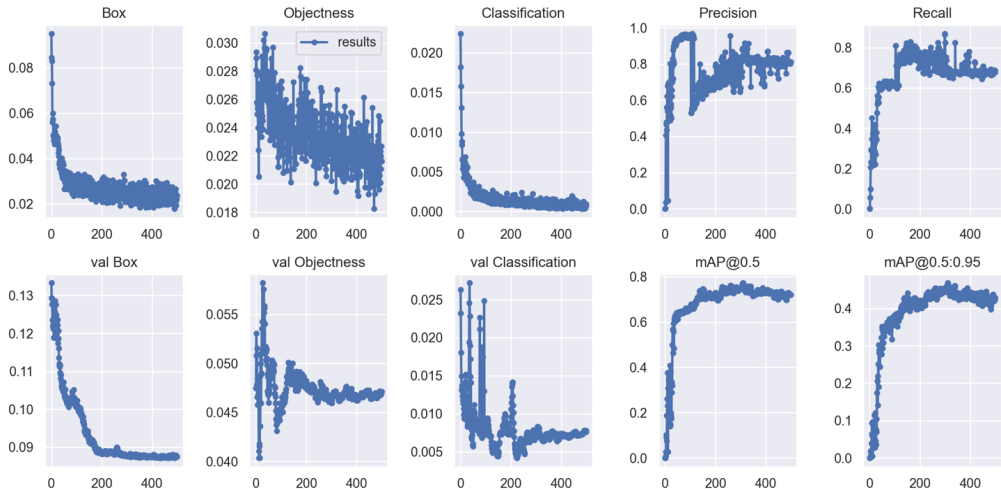

**Fig.21.** YOLOv7 training results

In the case of YOLOv8, the training utilized YOLOv8n initial weights, with an input size of 640*640 and a batch size of 4. Training was conducted across three phases: 5, 50, and 500 epochs. Training was automatically halted at 113 rounds during the 500-epoch phase. The results are illustrated in Fig. 23.

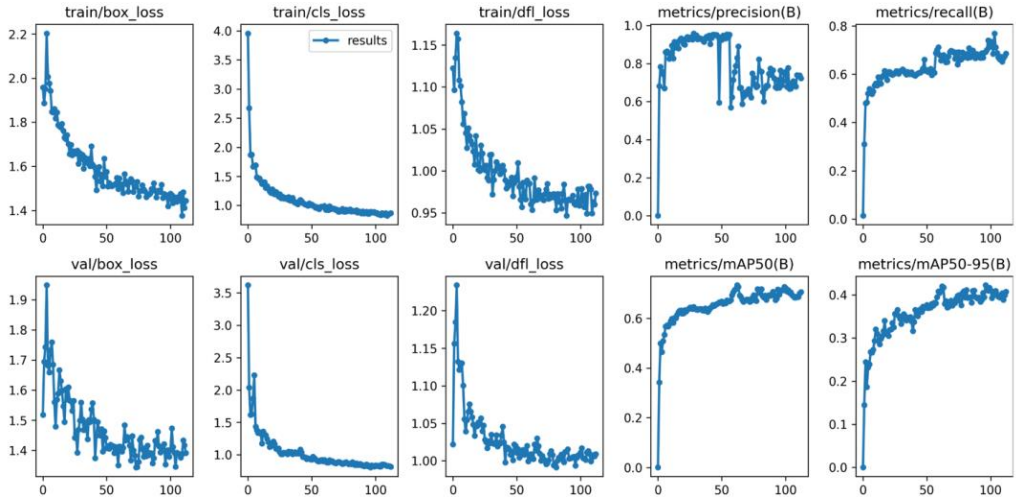

**Fig. 22.** YOLOv8 training results

As can be seen from the Fig. 20 to Fig. 22, the algorithm tends to stabilize the detection accuracy after iteration, and the errors fluctuate at a stable value, indicating that the detection effect is optimal at this time.

The YOLO object detection model, trained using the PyTorch framework, exhibits high performance and accuracy. Once the training is completed, the pre-trained model is converted to the mlmodel format using coremltools, a Python library, thereby broadening its application scope. Subsequently, the converted model is deployed on Apple devices, fully harnessing the hardware acceleration capabilities of these devices.

*5.3 Feasibility experiments for object detection*

By testing the performance of the trained model and verifying whether the model can meet the requirements of real-time, this section mainly introduces the effect experiment of single-frame object detection and the effect experiment of continuous frame object detection. The test consists of two parts: a single image detection effect test and a sequential video frame object detection performance test. The test environment is shown in Table 4.

Table 4 Object detection effect test environment

| Configuration | Parameter |
| --- | --- |
| System | macOS 13.3 |
| RAM | 16GB |
| Chip | Apple M1 |
| NPU performance | 16core-11TOPS |

(1) Single-image object detection performance

The trained model was loaded using the Create ML App, and its recognition performance for a single frame was assessed, with the test results displayed in Fig. 23.

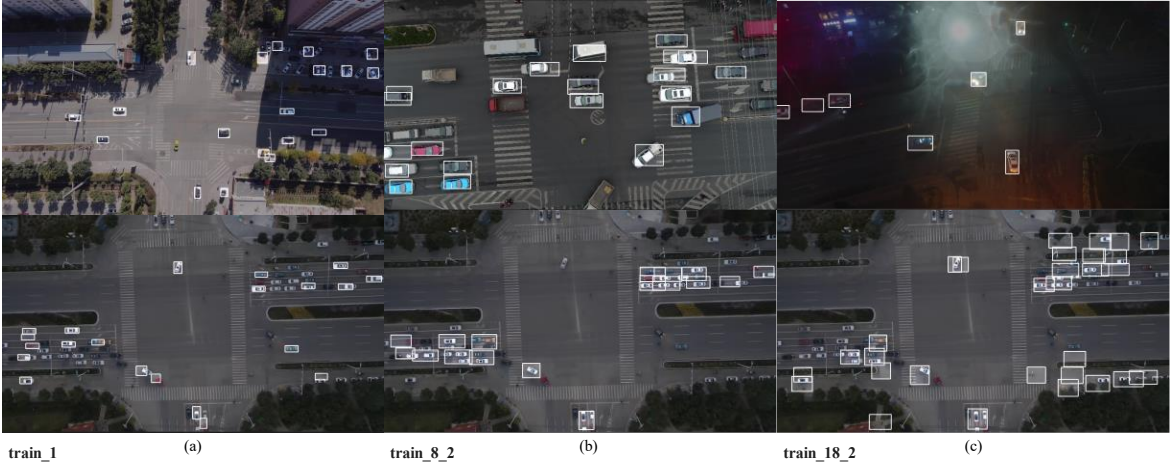

train_1 (a)     train_8_2 (b)     train_18_2 (c)

**Fig. 23.** Native model object detect effect

The analysis of test results indicates that under daylight conditions and in scenarios with fewer vehicles, the native model trained using Create ML proficiently recognizes the majority of vehicles. However, as the number of vehicles increases, the model's recognition rate diminishes, particularly in nocturnal settings. Notably, when confronted with targets of varying scales, the model's performance degrades, resulting in potential misalignment between the target box and the actual object. These challenges can be largely attributed to the inherent YOLOv2 algorithm integrated into the framework.

In comparison to the native model, as depicted in Fig. 24, it becomes evident that object detection performance, underpinned by the YOLO model, undergoes significant enhancement, resulting in markedly heightened accuracy. The YOLO model excels in identifying the majority of vehicle targets.

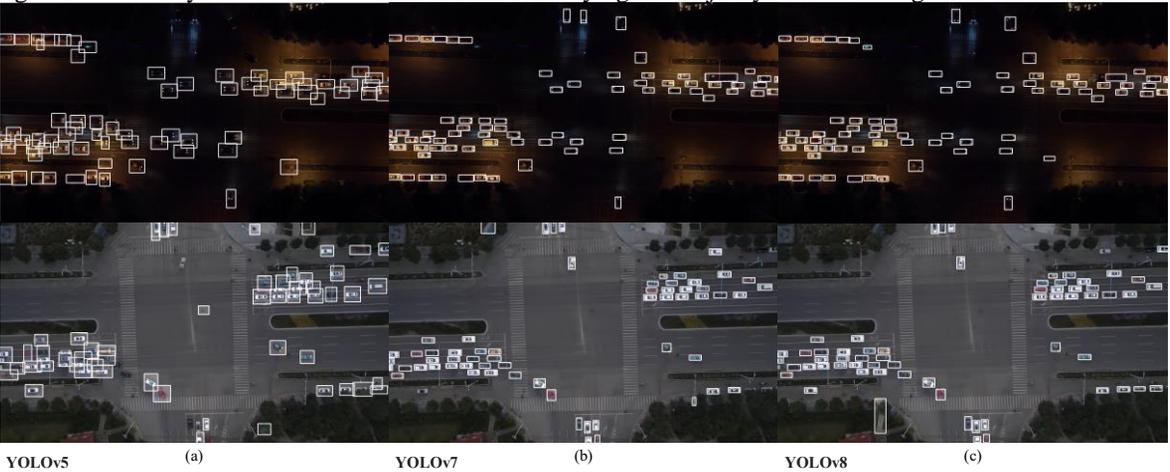

YOLOv5 (a)     YOLOv7 (b)     YOLOv8 (c)

**Fig. 24.** Third-party model object detect effect

(2) Sequential video frame object detection performance test

The system employed in this study, executed within a simulator environment, facilitates video frame analysis by employing the local album function of the system. The chosen test video is read and its individual frames are sequentially extracted. The ensuing test outcomes are presented in Table 5.

Table 5 Detection performance test results for each model

| Model | Size(MB) | Probe rate (FPS) | Meet real-time detection requirements |
|---|---|---|---|
| train_1 | 31.8 | 200+ | Yes |
| train_8_2 | 31.8 | 200+ | Yes |
| train_18_2 | 31.8 | 200+ | Yes |
| YOLOv5 | 83.9 | 60+ | Yes |
| YOLOv7 | 146.3 | 30+ | Yes |
| YOLOv8 | 12.1 | 140+ | Yes |

Based on the experimental findings, the object detection performance of each model is delineated as follows:

1) Native models: The three native models, namely, train_1, train_8_2, and train_18_2, demonstrate uniform characteristics concerning their size and sheer detection rate, with each model's dimensions measuring 31.8MB and achieving over 200 frames per second (FPS). This illustrates that, following optimization within the system, native models manage to attain a high detection rate while maintaining a compact file size, thereby aligning with the requisites of real-time object detection. Nonetheless, it should be noted that these native models exhibit a suboptimal target recognition rate and may not be well-suited for handling complex real-world scenarios.

2) YOLOv5: In contrast to native models, the YOLOv5 model experiences a nearly threefold increase in size, reaching 83.9MB, while its pure detection rate decreases to 60+ FPS. Despite this reduction in detection speed, YOLOv5 continues to fulfill the performance requirements for real-time object detection. Furthermore, YOLOv5 outperforms native models in terms of target recognition, rendering it more favorable for practical applications.

3) YOLOv7: The YOLOv7 model exhibits the largest size, measuring 146.3MB, which is nearly five times the size of the original model. Simultaneously, the pure detection rate remains above 30 FPS, essentially meeting the performance prerequisites for real-time target detection.

4) YOLOv8: The YOLOv8 model excels in terms of file size, with a compact 12.1MB, nearly three times smaller than the native model. Although its pure detection rate (140+ FPS) is slightly lower than that of the native model, it maintains strong real-time performance. Additionally, YOLOv8 enhances the target recognition rate, establishing its robust competitiveness in practical application scenarios.

Considering the size, detection rate, and target recognition rate of each model, this study draws the following conclusions: while native models offer advantages in size and speed, their target recognition rate remains modest. In contrast, third-party YOLO series models boast high target recognition rates while preserving real-time performance, rendering them better suited for practical application scenarios. Consequently, this study recommends prioritizing the utilization of YOLO series models in practical projects to meet performance and recognition rate requirements.

*5.4 Target tracking performance testing Actual flight experiments*

Through the system's functionality enabling the reading of local albums, the test video is systematically processed, and video frames are subsequently extracted in a sequential manner. The refined tracking mechanisms are vividly exemplified in Fig. 25. In subfigure (a), the tracking effectiveness is demonstrated as id39 executes a left turn from north to east, and id29 proceeds straight from north to south at frames 112, 274, and 439, leveraging the YOLOv7 model. Notably, at frame 274, id39 is discerned as executing a left turn. Subfigure (b) displays the tracking outcomes as id27 travels straight from east to west while id7 advances straight from west to east at frames 1235, 1458, and 1669, capitalizing on the YOLOv8 model. This approach successfully identifies id27's left lane change behavior at frame 1458.

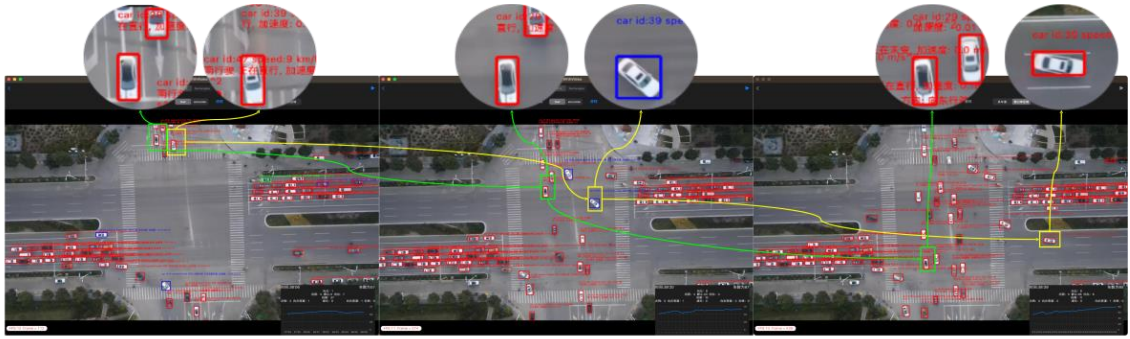

(a)

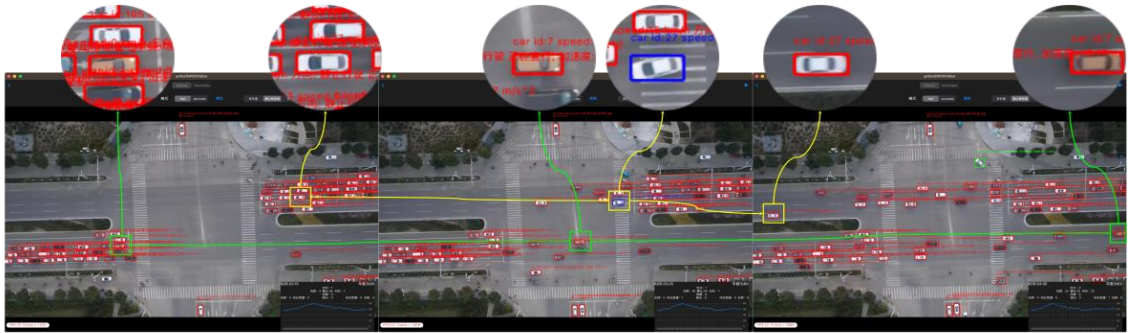

(b)

**Fig. 25.** Improved tracking effect

The comprehensive test results are elucidated in Table 5.

Table 6 Improved tracking performance test results for each model

| Model | Size(MB) | Minimum tracking rate (FPS) | Meet real-time detection requirements |
|---|---|---|---|
| YOLOv7 | 146.3 | 11+ | Met in the case of dropped frames |
| YOLOv8 | 12.1 | 23+ | Met in the case of dropped frames |

Upon scrutinizing the experimental outcomes pertaining to the test video, it becomes evident that the implementation of the SORT algorithm has markedly enhanced target tracking performance. In direct comparison to the Vision tracker's tracking rate, the YOLOv7 model sustains a commendable tracking rate of approximately 11 frames per second (FPS) even when the scene features a high density of vehicles. Though it surpasses YOLOv8 in terms of recognition effectiveness, the YOLOv7 model lags in real-time processing capacity. In contrast, the refined YOLOv8 model exhibits superior stability and precision, maintaining a tracking rate of roughly 23 FPS under high-vehicle-density scenarios. As the vehicle count diminishes, both models comfortably attain rates exceeding 30 FPS, adequately meeting real-time tracking requirements. This underscores the SORT algorithm's dual role in not only enhancing tracking efficiency but also ensuring superior tracking precision.

*5.5 Actual flight experiments*

In the conducted real-flight experiment, the study focused on multiple intersections located in Hohhot, Inner Mongolia. The drone operated at altitudes ranging from 80 to 120 meters, encountering clear weather conditions, mild wind forces ranging from 2 to 4, and conducted flights during the time frame of 1 to 5 pm. The flight equipment employed was the DJI Mavic Mini drone, and the mobile device utilized was the iPad Pro 2020.

Throughout the real-flight experiments, certain targets exhibited challenges in identification, resulting in either incorrect identifications or missed identifications. By systematically analyzing the experimental video footage captured during the actual flight experiment, these occurrences were categorized and quantified at 15-second intervals. The analysis encompassed a total duration of 62 minutes and 39 seconds. The assessment outcomes are comprehensively presented in Table 7.

Table 7 Performance Evaluation Metrics

| TP | FP | FN | Precision(%) | Recall(%) | F1 (%) |
|---|---|---|---|---|---|
| 2956 | 52 | 406 | 98.27 | 87.93 | 92.85 |

Detection target plotting and data visualization are achieved through two distinct views. The result plotting module is responsible for receiving data from the data processing module and rendering the vehicle target frames along with the corresponding micro traffic data. On the other hand, the data visualization interface extracts macro traffic data from the database and presents it in a visual format. As depicted in Fig.26, the user can seamlessly switch between the a and b views by clicking on the corresponding labeled area. During the real flight experiment, clicking on this area triggers the swift transition between views (a) and (b).

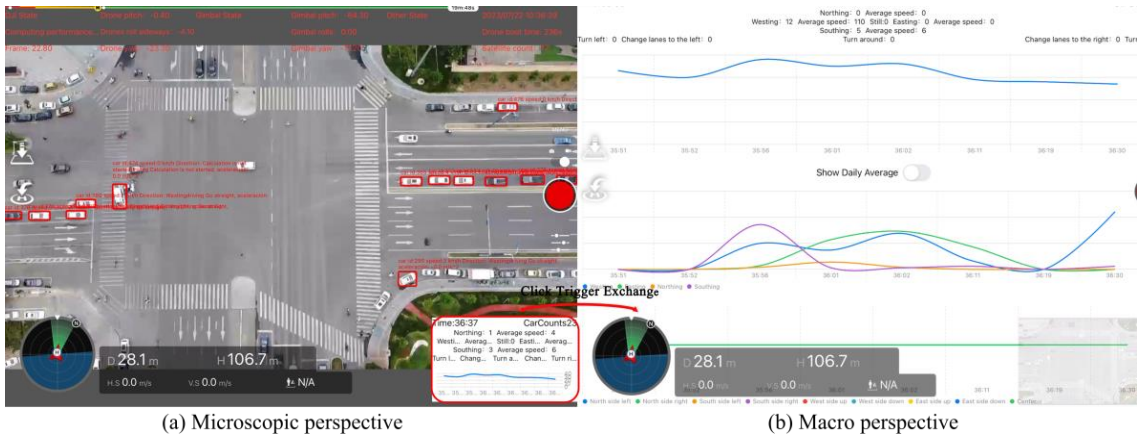

(a) Microscopic perspective　　　　　　　　　　(b) Macro perspective

**Fig. 26.** Micro view and macro view switching diagram

From a micro view, the system provides valuable insights into individual vehicles' acceleration, speed, and direction of movement. The thumbnail view located at the bottom right allows users to observe the real-time count of vehicles within the UAV camera's field of view and the number of vehicles traveling in each direction. Moreover, by accurately calculating the speed of the detected vehicles, the system can also derive the average speed of vehicles moving in each specific direction. This comprehensive data visualization empowers users with a deeper understanding of the traffic dynamics and patterns in the monitored area.

From a macroscopic view, the system also provides statistics of the aforementioned microscopic parameters. Over the observation time of the drone, users can gain valuable insights into the fluctuation patterns of the number of vehicles and the average speed of vehicles in each direction. Additionally, the system can derive the changing trends in the average speed of each lane. This crucial information serves as a guide to identify congested lanes and directions in real-time, empowering users with the ability to make informed decisions and effectively manage traffic flow in congested areas.

# 6. Conclusions

This paper constructs a reliable and efficient real-time mobile UAV video-based vehicle data acquisition and analysis system. Main findings are summarized as followings:

1. This research presents a comprehensive solution that synergizes UAV development and deep learning technology on mobile terminals. Leveraging Swift and Objective-C languages, the iOS SDK serves as the fundamental framework for the program, while DJI Mobile SDK and UX SDK are utilized to develop UAV functionalities. Additionally, CoreML and the Vision framework are employed to deploy object detection models, culminating in the design of a real-time object detection and tracking system on iOS devices.

2. Through a meticulous comparison between the Create ML-trained native model and the YOLOv8 model, it is evident that the YOLO model surpasses the Create ML-trained native model in performance. To further enhance the tracking algorithm, the original Vision framework tracker is replaced with the SORT algorithm-based tracking module. By integrating the YOLOv8 model with the SORT algorithm, real flight tests are conducted using an iPad equipped with A12Z and a DJI Mavic Mini. The results show an impressive accuracy rate of 98.27% and a recall rate of 87.93%. when the number of vehicles reaches 27, the real-time processing power of the A12Z chip is lower than the drone video transmission rate. Therefore, this paper optimizes performance through a frame dropping strategy, selectively skipping non-critical video frames to enhance system efficiency without compromising accuracy and stability in the detection process.

3. Factors such as UAV camera parameters, flight parameters, and equipment hardware performance, are considered to dynamically calculate the FPS parameters for real-time video frame processing capability. This enables the acquisition of real-time micro parameters such as speed, acceleration, direction, and lane-changing behavior, along with macro parameters like the number of vehicles, total vehicles, and vehicles in all directions within the regional road section. The visualization of this traffic data is presented in real time, offering valuable insights into the traffic scenario.

This paper has made significant strides in the domain of real-time target detection and tracking for mobile UAVs. However, certain aspects demand further investigation and refinement. One crucial aspect pertains to the communication delay between UAVs and mobile devices, as it can impact the efficacy of real-time target detection and tracking. Addressing this challenge through communication-related optimizations will contribute valuable advancements to the system's overall efficiency and effectiveness.

Moreover, specific application scenarios, such as inclement weather conditions like rain and snow, as well as low-light environments, present unique challenges for detection and tracking performance. By devising targeted optimizations tailored to these scenarios, we can bolster the system's capabilities and ensure robust performance under diverse real-world conditions.

In conclusion, conducting further research to address communication delays and exploring specialized optimizations for specific scenarios holds the potential to elevate real-time target detection and tracking for mobile UAVs. These advancements will enhance the system's practicality and reliability, making it a more valuable tool for a wide range of real-world applications.